\documentclass[letterpaper, 10 pt, conference]{ieeeconf}

\IEEEoverridecommandlockouts
\usepackage{cite}
\usepackage{multicol}
\usepackage[bookmarks=true]{hyperref}
\usepackage{amsmath}
\usepackage{amssymb}
\usepackage[ruled,vlined]{algorithm2e}
\usepackage{todonotes}
\usepackage{tikz}
\usepackage{pgfplots}
\pgfplotsset{compat=1.18}
\usetikzlibrary{arrows.meta, positioning, calc, backgrounds, fit}

\begin{document}

\title{\LARGE \bf
    World Translation:
    Minimizing Sim-to-Real Gap with Backward Dynamics Extraction and Unpaired Domain Translation
}

\author{Xinchen Yao$^{1}$, Leixing Chang$^{1}$, and Hua Chen$^{2}$
\thanks{$^{1}$ Zhejiang University}
\thanks{$^{2}$ LimX Dynamics}
}

\maketitle
\thispagestyle{empty}
\pagestyle{empty}

\begin{abstract}
    The gap between simulation and reality remains a fundamental challenge in deploying simulation-trained robotic policies in the real world. 
    Real-to-sim methods narrow this gap from the real side, learning transition dynamics from real data to build a more realistic digital world.
    Learned dynamics models are their dominant instance. Such methods, however, face a partial observability problem: the same observation may branch to different transitions due to unobservable factors.
    Existing methods assume these factors can be recovered from observation history.  However, this may fail whenever observation history is uninformative, such as a sudden contact event with no prior warning.
    To address this limitation, we propose \textit{World Translation}, which exploits a complementary strength of simulators and learned dynamics. Simulators are deterministic but physically imperfect, while learned models are accurate but underdetermined under partial observability.
    Rather than predicting transitions forward from history, we extract the unobservable dynamics information backward from an observed transition, then translate this feature across simulation and reality as an unpaired domain-translation problem that preserves dynamics content while transferring domain style.
    Experiments across humanoid, quadruped, and manipulator platforms show that our method achieves more accurate dynamics modeling than baselines, with the largest gains when unobservable factors cannot be recovered from observation history. Real-robot deployment on Go2 quadruped confirms improved policy transfer.
    
\end{abstract}

\section{Introduction} \label{sec:intro}
Sim-to-real transfer is fundamental to robot learning, yet the gap between simulation and reality remains a challenge~\cite{hwangbo2019learning,lee2020learning,singh2024dextrah}.
The two dominant mitigation strategies, domain randomization (DR)~\cite{tobin2017domain,peng2018sim,akkaya2019solving} and system identification (SysID)~\cite{ljung1998system,chebotar2019closing}, each carries a structural cost.
DR buys robustness by sacrificing policy optimality~\cite{josifovski2022analysis,xiao2025learning}, while SysID calibrates a predefined model and therefore cannot account for unmodeled dynamics.
Learned dynamics models~\cite{hafner2019learning,hafner2020dream,hafner2023mastering,hansen2023td}, including residual variants~\cite{golemo2018sim,heiden2021neuralsim,sontakke2023residual}, take a more promising route, fitting transition dynamics directly from real data and can in principle capture the very phenomena the other two miss.
That promise, however, is undercut by partial observability. The same observation-action pair can branch to different transitions because of unobserved factors. Existing models assume these are recoverable from observation history, which fails whenever the history is uninformative, such as a sudden contact in proprioceptive locomotion, forcing the prediction to average over the possible transition modes.

\begin{figure}[t]
    \centering
    \includegraphics[width=\linewidth]{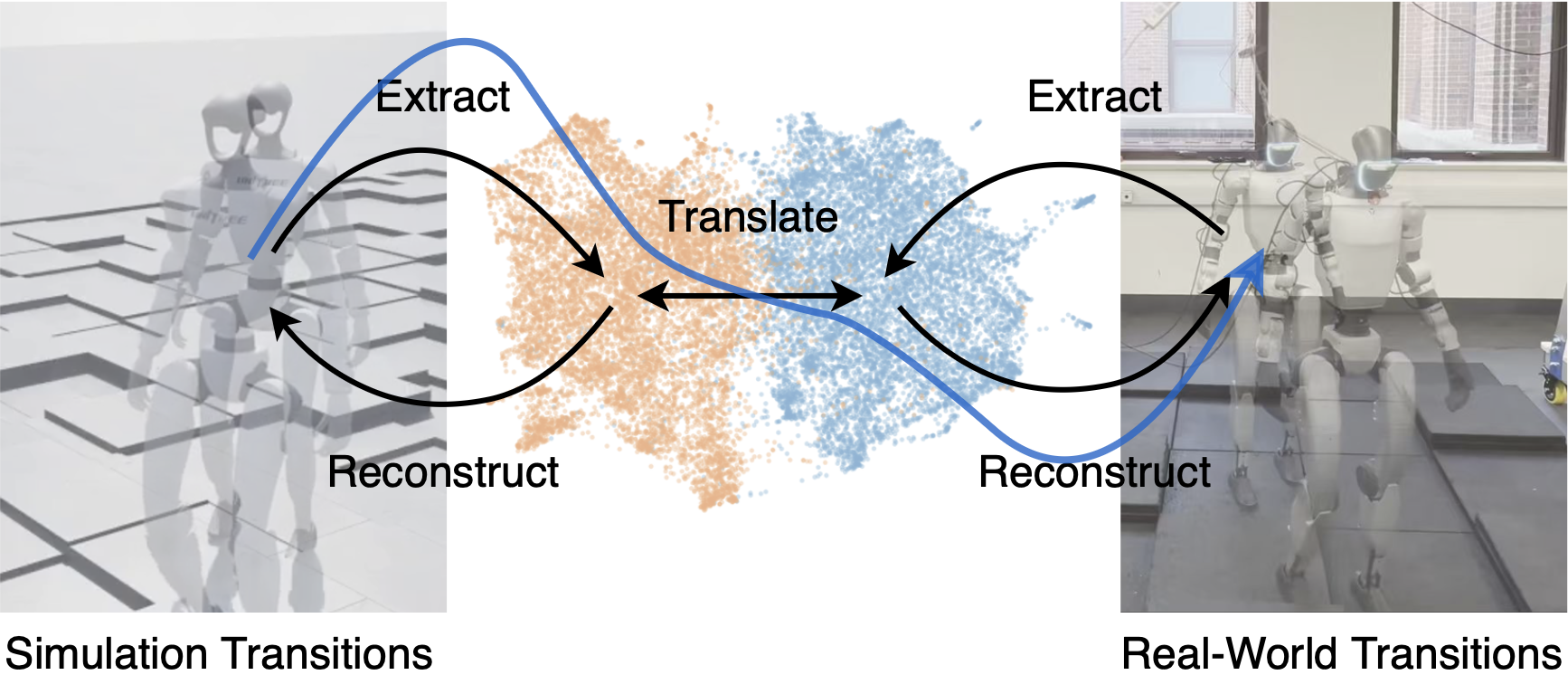}
    \caption{World Translation bridges the sim-to-real gap by extracting dynamics features from observed transitions and translating them across domains. Black arrows are used during training; the blue arrow illustrates the inference pipeline, where simulation dynamics features are translated and reconstructed to predict real-world transitions.}
    \label{fig:cover}
    \vspace{-20pt}
\end{figure}
Our key observation is that simulators and learned models fail in opposite ways. A simulator keeps a complete internal state and predicts deterministically but with physically inaccurate transitions, whereas a learned model predicts accurately from data but stays underdetermined under partial observability. The signal the learned model lacks is not gone. It lives in the transition outcome, which a simulator can observe at will. This motivates \textit{backward dynamics extraction}. Rather than predicting a transition forward from history, we encode the unobservable dynamics information backward from an already-observed transition, directly capturing what causes transition uncertainty. Because the same hidden factor leaves different signatures across domains, features extracted in one cannot be used directly in the other. We address this with \textit{unpaired domain translation}, which maps these features between simulation and reality, using cycle consistency to preserve dynamics content while transferring domain style. We call the resulting framework \textbf{World Translation}.

Our contributions can be summarized as follows: 
    \textbf{1) Backward dynamics extraction:} unobservable dynamics information is recoverable from transition outcomes even when observation history is insufficient to recover it, making backward extraction more broadly reliable than history-based methods.
    \textbf{2) Unpaired domain translation over dynamics features:} cycle consistency over extracted dynamics features enables domain transfer without requiring paired sim-real transitions.
    \textbf{3) Empirical validation} on humanoid, quadruped, and manipulator platforms, with real-robot Go2 deployment showing improved policy performance.

\section{Related Work}

\subsection{Sim-to-Real and Learned Dynamics}
Sim-to-real transfer is commonly addressed through domain randomization~\cite{tobin2017domain,peng2018sim,akkaya2019solving,tan2018sim} and system identification~\cite{ljung1998system,chebotar2019closing}, with recent hybrids~\cite{tiboni2023dropo,mehta2020active}. As noted in Section \ref{sec:intro}, both trade away either policy optimality or the ability to model phenomena outside a predefined structure. Learned dynamics models instead build a realistic physics environment directly from real transition data. PILCO~\cite{deisenroth2011pilco} showed sample efficiency via Gaussian processes and PETS~\cite{chua2018deep} extended this to neural networks. The Dreamer family~\cite{hafner2019learning,hafner2020dream,hafner2023mastering} learns latent dynamics from high-dimensional observations. DayDreamer~\cite{wu2023daydreamer} adapts online on real robots.
TD-MPC2~\cite{hansen2023td} couples world models with model predictive control. Despite their differences, most assume the unobservable factors are recoverable from observation history.

\subsection{Residual Dynamics and Action}

Residual methods aim to produce more realistic next observations by augmenting the simulator with learned corrections: residual dynamics methods correct outputs directly~\cite{golemo2018sim,heiden2021neuralsim,gao2024sim,sontakke2023residual}, while residual action methods modify inputs so the simulator naturally produces better outcomes~\cite{hanna2017grounded,karnan2020rgat,he2025asap}. Both treat the simulator as a fixed prior to be corrected, rather than a source of dynamics information, and share this history-recoverability assumption.

\subsection{Unpaired Domain Translation}

CycleGAN~\cite{zhu2017unpaired} introduced cycle-consistent adversarial training for unpaired image-to-image translation. It has been applied to visual sim-to-real transfer~\cite{rao2020rl,ho2021retinagan}, addressing only the visual gap. We apply unpaired translation to learned dynamics features rather than images, enabling simulation to supply dynamics information that forward-prediction objectives may not recover from observation history.

\section{Problem Formulation}
\label{sec:problem}

\subsection{System Dynamics and Hidden Variables}

We now formalize the problem by introducing two types of unobservable factors: time-varying \textit{hidden variables} $\mathbf{h}_t$ and time-invariant \textit{domain characteristics} $\mathbf{c}$.
Consider a robotic system with inaccessible full state $\mathbf{s}_t \in \mathcal{S}$, accessible observation $\mathbf{o}_t \in \mathcal{O}$, and accessible action $\mathbf{a}_t \in \mathcal{A}$. The dynamics evolution is deterministic in state space, and we have $\mathbf{s}_{t+1} = g(\mathbf{s}_t, \mathbf{a}_t)$. However, due to the information gap between $\mathbf{s}_t$ and $\mathbf{o}_t$, identical observation-action pairs can produce different outcomes, and the mapping $(\mathbf{o}_t, \mathbf{a}_t) \mapsto \mathbf{o}_{t+1}$ is not deterministic.
To model the unobservable factors and restore determinism, we introduce \textit{hidden variables} $\mathbf{h}_t \in \mathcal{H}$, capturing the unobservable component of the full state; $\mathbf{h}_t$ may vary unpredictably. By conditioning on $\mathbf{h}_t$, the dynamics become deterministic:
\begin{equation}
    \mathbf{o}_{t+1} = f(\mathbf{o}_t, \mathbf{a}_t, \mathbf{h}_t).
    \label{eq:hidden-dynamics}
\end{equation}
We make no assumptions about $p(\mathbf{h}_{t+1}|\mathbf{h}_t,\mathbf{o}_{t},\mathbf{a}_t)$, requiring only that the effect of $\mathbf{h}_t$ be identifiable from the observed transition $(\mathbf{o}_t, \mathbf{a}_t) \to \mathbf{o}_{t+1}$.

\subsection{Backward Dynamics Extraction}
\label{sec:backward}

As a learned dynamics model, our goal is to predict $\mathbf{o}_{t+1}$. As formulated in \eqref{eq:hidden-dynamics}, this requires knowledge of hidden variables $\mathbf{h}_t$. While $\mathbf{h}_t$ cannot be directly observed, its effect on dynamics is detectable through state transitions. We introduce a \textit{dynamics feature} $\mathbf{z}_t = \omega(\mathbf{h}_t)$ to represent this information in latent space.
\begin{align}
    (\mathbf{o}_t, \mathbf{a}_t, \mathbf{o}_{t+1}) & \xrightarrow{\;\text{encode}\;} \mathbf{z}_t = \omega(\mathbf{h}_t), \\
    (\mathbf{o}_t, \mathbf{a}_t, \mathbf{z}_t)     & \xrightarrow{\;\text{decode}\;} \mathbf{o}_{t+1}.
\end{align}
We call this backward extraction because it uses future information $\mathbf{o}_{t+1}$ to infer $\mathbf{z}_t$ at time $t$.

However, this extraction alone does not enable dynamics prediction. $\mathbf{z}_t$ requires observing the outcome $\mathbf{o}_{t+1}$, which is precisely what we wish to predict. 
To make $\mathbf{z}_t$ useful, we need an alternative source, such as a simulator where transitions \textit{can} be observed, from which we extract dynamics features and translate them to the target system.

\subsection{Unpaired Domain Translation}
\label{sec:unpaired}

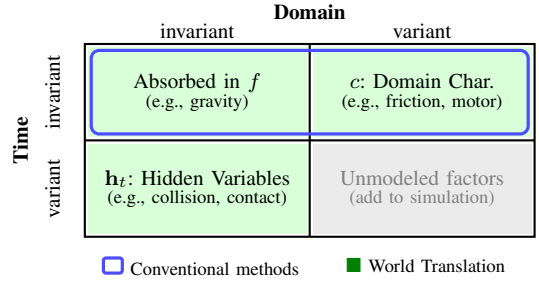
\begin{figure}[t]
    \centering
    \begin{tikzpicture}[scale=0.85]
    \fill[green!15] (0.05,1.55) rectangle (3.45,2.95);
    \fill[green!15] (3.55,1.55) rectangle (6.95,2.95);
    \fill[green!15] (0.05,0.05) rectangle (3.45,1.45);
    \fill[black!8] (3.55,0.05) rectangle (6.95,1.45);

    \draw[thick] (0,0) rectangle (7,3);
    \draw[thick] (3.5,0) -- (3.5,3);
    \draw[thick] (0,1.5) -- (7,1.5);

    \draw[blue!70, very thick, rounded corners=3pt] (0.1,1.6) rectangle (6.9,2.9);

    \node[font=\footnotesize\bfseries] at (3.5,3.5) {Domain};
    \node[font=\footnotesize] at (1.75,3.2) {invariant};
    \node[font=\footnotesize] at (5.25,3.2) {variant};

    \node[font=\footnotesize\bfseries, rotate=90] at (-1.0,1.5) {Time};
    \node[font=\footnotesize, rotate=90] at (-0.5,2.25) {invariant};
    \node[font=\footnotesize, rotate=90] at (-0.5,0.75) {variant};

    \node[align=center, font=\footnotesize] at (1.75,2.25) {Absorbed in $f$\\[-2pt]{\scriptsize(e.g., gravity)}};

    \node[align=center, font=\footnotesize] at (5.25,2.25) {$c$: Domain Char.\\[-2pt]{\scriptsize(e.g., friction, motor)}};

    \node[align=center, font=\footnotesize] at (1.75,0.75) {$\mathbf{h}_t$: Hidden Variables\\[-2pt]{\scriptsize(e.g., collision, contact)}};

    \node[align=center, font=\footnotesize, text=black!50] at (5.25,0.75) {Unmodeled factors\\[-2pt]{\scriptsize(add to simulation)}};

    \node[font=\scriptsize] at (1.8,-0.45) {\tikz\draw[blue!70, very thick, rounded corners=1pt] (0,0) rectangle (0.25,0.2); Conventional methods};
    \node[font=\scriptsize] at (5.3,-0.45) {\textcolor{green!50!black}{$\blacksquare$} World Translation};
\end{tikzpicture}
    \caption{Taxonomy of unobservable factors. Conventional dynamics models implicitly learn time-invariant domain characteristics ($\mathbf{c}$) but cannot explicitly capture time-variant hidden variables ($\mathbf{h}_t$) when history is insufficient to recover them. World Translation captures $\mathbf{h}_t$ through backward extraction, enabling generalization across cases where history-based inference fails. Factors that are both domain-variant and time-variant (e.g., a payload present in reality but absent in simulation) represent unmodeled phenomena that World Translation cannot recover from the source system.}
    \label{fig:factors}
    \vspace{-15pt}
\end{figure}

The backward extraction in Section~\ref{sec:backward} assumes a single domain where $\mathbf{z}_t = \omega(\mathbf{h}_t)$. However, when applying this framework across domains, for example, simulation and reality, the same hidden variable $\mathbf{h}_t$ produces different observed transitions in each domain. Therefore, the dynamics feature extracted from transitions is not solely a function of $\mathbf{h}_t$, but also depends on which domain generated the transition.
We introduce \textit{domain characteristics} $\mathbf{c}$ to explicitly capture these systematic differences. With $\mathbf{h}_t$ and $\mathbf{c}$, the dynamics can be formulated as $\mathbf{o}_{t+1} = f(\mathbf{o}_t, \mathbf{a}_t, \mathbf{h}_t, \mathbf{c})$. $\mathbf{c}$ is an implicitly learned representation capturing domain-specific effects, including unmodeled phenomena.
Figure~\ref{fig:factors} illustrates how $\mathbf{h}_t$ and $\mathbf{c}$ relate to other unobservable factors.

In a multi-domain setting, the dynamics feature $\mathbf{z}_t$ is determined by both hidden variables and domain characteristics:
\begin{align}
    (\mathbf{o}_t, \mathbf{a}_t, \mathbf{o}_{t+1}) & \xrightarrow{\;\text{encode}\;} \mathbf{z}_t = \omega(\mathbf{h}_t, \mathbf{c}), \\
    (\mathbf{o}_t, \mathbf{a}_t, \mathbf{z}_t)     & \xrightarrow{\;\text{decode}\;} \mathbf{o}_{t+1}.
\end{align}
For the same $\mathbf{h}_t$, $\mathbf{z}_t^A = \omega(\mathbf{h}_t, \mathbf{c}^A)$ and $\mathbf{z}_t^B = \omega(\mathbf{h}_t, \mathbf{c}^B)$. We seek $G_{A \rightarrow B}$ such that:
\begin{equation}
    G_{A \rightarrow B}(\omega(\mathbf{h}_t, \mathbf{c}^A)) \approx \omega(\mathbf{h}_t, \mathbf{c}^B).
\end{equation}
This is an \textit{unpaired} translation problem: transitions collected with matching $(\mathbf{o}_t, \mathbf{a}_t)$ across domains cannot guarantee identical underlying $\mathbf{h}_t$, as physical interactions cannot be precisely reproduced.

\subsection{Combined Objective}

World Translation learns a feature $\mathbf{z}_t$ and a cross-domain mapping $G_{A \rightarrow B}$, such that observing a transition in domain $A$ predicts the corresponding transition in domain $B$:
\begin{equation}
    \mathbf{o}_{t+1}^A \xrightarrow[\;\mathbf{o}_t, \mathbf{a}_t\;]{\;\text{encode}\;} \mathbf{z}_t^A \xrightarrow{\;G_{A \rightarrow B}\;} \mathbf{z}_t^B \xrightarrow[\;\mathbf{o}_t, \mathbf{a}_t\;]{\;\text{decode}\;} \mathbf{o}_{t+1}^B.
    \label{eq:pipeline}
\end{equation}

\section{Method}
\label{sec:method}

We instantiate the World Translation framework defined in Section~\ref{sec:problem} using a VAE with auxiliary regularization for backward dynamics extraction and a CycleGAN \cite{zhu2017unpaired} for unpaired domain translation. 

\subsection{Backward Dynamics Extraction}

We implement the dynamics encoder-decoder 
using a variational autoencoder. We first describe the core architecture, then analyze two challenges that arise during training and introduce auxiliary objectives to address them.

The encoder $E_\phi$ maps a transition tuple to a latent distribution $(\boldsymbol{\mu}, \log \boldsymbol{\sigma}^2) = E_\phi(\mathbf{o}_t, \mathbf{a}_t, \mathbf{o}_{t+1})$,
where the latent code is sampled as $\mathbf{z}_t \sim \mathcal{N}(\boldsymbol{\mu}, \text{diag}(\boldsymbol{\sigma}^2))$.
The decoder $D_\theta$ reconstructs the next observation: $\hat{\mathbf{o}}_{t+1} = D_\theta(\mathbf{o}_t, \mathbf{a}_t, \mathbf{z}_t)$.
Ideally, the encoder encodes both $\mathbf{h}_t$ and $\mathbf{c}$ into $\mathbf{z}_t$, but two challenges prevent this from happening naturally.

\textit{Challenge 1: Direct encoding of $\mathbf{o}_{t+1}$.} The encoder might bypass learning meaningful dynamics by directly encoding $\mathbf{o}_{t+1}$ into $\mathbf{z}_t$. We introduce a blind decoder $D_\psi$ that attempts to predict the next observation from $\mathbf{z}_t$ alone:
$\tilde{\mathbf{o}}_{t+1} = D_\psi(\mathbf{z}_t)$.
If the blind decoder can accurately predict $\mathbf{o}_{t+1}$ from $\mathbf{z}_t$ without access to $(\mathbf{o}_t, \mathbf{a}_t)$, then $\mathbf{z}_t$ contains direct observation information rather than abstract dynamics features.
The encoder minimizes reconstruction error while simultaneously \textit{maximizing} the blind decoder's prediction error. The blind decoder tries to minimize its own error, creating an adversarial objective:
\begin{equation}
    \mathcal{L}_\text{blind} = \lambda_b \exp\left(-\frac{\|\mathbf{o}_{t+1} - \tilde{\mathbf{o}}_{t+1}\|^2}{\sigma_b^2}\right)
\end{equation}
The exponential form provides bounded gradients and saturates smoothly when the blind decoder error is already high, avoiding unnecessary optimization pressure. The temperature $\sigma_b$ controls sensitivity. The blind decoder is trained separately with detached gradients from the encoder.

\textit{Challenge 2: Information leakage in $(\mathbf{o}_t, \mathbf{a}_t)$.} The data-collecting policy reacts to both $\mathbf{h}_t$ and $\mathbf{c}$, so $(\mathbf{o}_t, \mathbf{a}_t)$ carries partial information about both. For $\mathbf{h}_t$, this is benign: the encoder only needs to extract the remaining information into $\mathbf{z}_t$, and during translation $(\mathbf{o}_t, \mathbf{a}_t)$ stays unchanged, so complete $\mathbf{h}_t$ semantics are preserved. For $\mathbf{c}$, however, since the decoder can infer domain information from $(\mathbf{o}_t, \mathbf{a}_t)$ alone, the encoder has no incentive to encode $\mathbf{c}$ into $\mathbf{z}_t$, but translation requires $\mathbf{c}$ to be encoded in $\mathbf{z}_t$. We introduce two auxiliary components. First, we add a domain classifier $C_\xi$ that predicts which domain a latent code originates from: $\hat{d} = C_\xi(\mathbf{z}_t)$. By requiring the classifier to succeed, we force $\mathbf{z}_t$ to encode domain-specific information:
\begin{equation}
    \mathcal{L}_\text{cls} = \lambda_c \cdot \text{BCE}(C_\xi(\mathbf{z}_t), d),
\end{equation}
where $d \in \{0, 1\}$ is the domain label and BCE denotes binary cross-entropy.
Even with domain information in $\mathbf{z}_t$, the decoder might still infer $\mathbf{c}$ from $(\mathbf{o}_t, \mathbf{a}_t)$ and ignore the domain information encoded in $\mathbf{z}_t$. We replace the basic MLP decoder with Feature-wise Linear Modulation (FiLM)~\cite{perez2018film}, where $\mathbf{z}_t$ conditions each layer:
\begin{equation}
\begin{aligned}
    D_\theta(\mathbf{o}_t, \mathbf{a}_t, \mathbf{z}_t) & = \text{FiLM}(\mathbf{o}_t, \mathbf{a}_t; \mathbf{z}_t)                                                    \\
                                                       & = \boldsymbol{\gamma}(\mathbf{z}_t) \odot f(\mathbf{o}_t, \mathbf{a}_t) + \boldsymbol{\beta}(\mathbf{z}_t),
\end{aligned}
\end{equation}
where $\odot$ denotes element-wise multiplication. FiLM has an inductive bias toward using $\mathbf{z}_t$ as the dominant conditioning signal, discouraging the decoder from bypassing it even when $(\mathbf{o}_t, \mathbf{a}_t)$ provides partial domain cues.
While these auxiliary components do not theoretically guarantee the decoder uses domain information from $\mathbf{z}_t$, we find empirically that they are effective, as reported in Section~\ref{sec:domain-char-exp}.

The full training objective combines all components:
\begin{equation}
\begin{aligned}
    \mathcal{L}_\text{VAE} = \; & \underbrace{\|\mathbf{o}_{t+1} - \hat{\mathbf{o}}_{t+1}\|^2}_{\text{reconstruction}} + \underbrace{\beta \cdot \text{KL}\left(q_\phi(\mathbf{z}_t) \| p(\mathbf{z}_t)\right)}_{\text{regularization}}  \\
                                & + \underbrace{\lambda_b \exp\left(-\frac{\|\mathbf{o}_{t+1} - \tilde{\mathbf{o}}_{t+1}\|^2}{\sigma_b^2}\right)}_{\text{blind penalty}}                                                                          \\
                                & + \underbrace{\lambda_c \cdot \text{BCE}(C_\xi(\mathbf{z}_t), d)}_{\text{domain classification}}
\end{aligned} \label{eq:vae_loss}
\end{equation}

\subsection{Unpaired Domain Translation}

Between simulation domain $S$ and real-world domain $R$, we implement domain translation using CycleGAN~\cite{zhu2017unpaired} with generators $G_{S \to R}$ and $G_{R \to S}$ (residual MLPs) and discriminators $D_S$ and $D_R$. The generators are trained with standard adversarial loss, cycle consistency loss (ensuring $\mathbf{z}^S \approx G_{R \to S}(G_{S \to R}(\mathbf{z}^S))$), and optional identity loss. This preserves hidden variable semantics in $\mathbf{z}_t$ while transforming domain characteristics.

\subsection{Training and Deployment Pipeline}

Algorithm~\ref{alg:world-translation-training} summarizes the training procedure. We train the VAE and CycleGAN jointly, with encoder-decoder shared across both domains. This shared representation ensures that dynamics features from different domains occupy a common latent space, making translation meaningful.

\begin{algorithm}[t]
    \small
    \caption{World Translation Training}
    \label{alg:world-translation-training}
    \DontPrintSemicolon
    \SetAlgoLined
    Initialize encoder $E_\phi$, decoder $D_\theta$, blind decoder $D_\psi$, classifier $C_\xi$. Initialize generators $G_{S \to R}, G_{R \to S}$ and discriminators $D_S, D_R$\;
    \While{not converged}{
        \tcp*[h]{\textit{Backward Dynamics Extraction}}
        Sample transitions $\{(\mathbf{o}_t, \mathbf{a}_t, \mathbf{o}_{t+1})\} \sim \mathcal{D}_S \cup \mathcal{D}_R$\;
        Encode: $\mathbf{z}_t \sim q_\phi(\mathbf{z}_t \mid \mathbf{o}_t, \mathbf{a}_t, \mathbf{o}_{t+1})$\;
        Decode: $\hat{\mathbf{o}}_{t+1} = D_\theta(\mathbf{o}_t, \mathbf{a}_t, \mathbf{z}_t)$\;
        Blind predict: $\tilde{\mathbf{o}}_{t+1} = D_\psi(\mathbf{z}_t)$\;
        Classify: $\hat{d} = C_\xi(\mathbf{z}_t)$\;
        Update $\phi, \theta, \xi$: minimize $\mathcal{L}_\text{VAE}$\;
        Update $\psi$: minimize $\|\mathbf{o}_{t+1} - \tilde{\mathbf{o}}_{t+1}\|^2$ with $\mathbf{z}_t$ detached\;

        \tcp*[h]{\textit{Unpaired Domain Translation}}
        Sample domain-separated batches from $\mathcal{D}_S$ and $\mathcal{D}_R$\;
        Encode: $\mathbf{z}^S \sim q_\phi(\mathbf{z} \mid \mathcal{D}_S)$, $\mathbf{z}^R \sim q_\phi(\mathbf{z} \mid \mathcal{D}_R)$\;
        Translate: $\hat{\mathbf{z}}^R = G_{S \to R}(\mathbf{z}^S)$, $\hat{\mathbf{z}}^S = G_{R \to S}(\mathbf{z}^R)$\;
        Cycle: $\tilde{\mathbf{z}}^S = G_{R \to S}(\hat{\mathbf{z}}^R)$, $\tilde{\mathbf{z}}^R = G_{S \to R}(\hat{\mathbf{z}}^S)$\;
        Update $G_{S \to R}, G_{R \to S}$: minimize $\mathcal{L}_\text{adv} + \lambda_\text{cyc}\mathcal{L}_\text{cyc} + \lambda_\text{id}\mathcal{L}_\text{id}$\;
        Update $D_S$: minimize $-\mathbb{E}[\log D_S(\mathbf{z}^S)] - \mathbb{E}[\log(1 - D_S(\hat{\mathbf{z}}^S))]$\;
        Update $D_R$: minimize $-\mathbb{E}[\log D_R(\mathbf{z}^R)] - \mathbb{E}[\log(1 - D_R(\hat{\mathbf{z}}^R))]$\;
    }
\end{algorithm}

At deployment, we apply the trained models during simulation. Given a simulator transition $(\mathbf{o}_t, \mathbf{a}_t, \mathbf{o}_{t+1}^\text{S})$, we encode, translate, and decode to obtain the aligned observation $\mathbf{o}_{t+1}^{S \to R}$, then \textit{overwrite} the simulator state rather than merely transforming observations. This ensures the policy trains on trajectories that evolve according to translated dynamics. Domain randomization is disabled during World Translation training to ensure deterministic simulation dynamics.

\section{Experiments}
\label{sec:experiments}

We evaluate World Translation across humanoid, quadruped, and manipulator platforms, asking: \textbf{1)} Can World Translation accurately predict dynamics, especially under unpredictable hidden variable evolution? \textbf{2)} Does the learned representation capture meaningful structure about hidden variables and domains? \textbf{3)} What components of our method are essential? \textbf{4)} Does our improved dynamics modeling lead to better downstream policy transfer?

\subsection{Dynamics Modeling Accuracy}

\subsubsection{Baselines}
For each platform, we create source and target domains in Isaac Lab with different physics parameters, such as stiffness, damping, mass and friction, to simulate the sim-to-real domain gap for more controllable evaluation. Transition data is collected using fixed pre-trained policies. The baselines are as follows.
\begin{itemize}
    \item Direct Prediction (DirectPred): Predicts from $(\mathbf{o}_t, \mathbf{a}_t)$ alone; error reflects hidden variable influence $\mathbf{h}_t$.
    \item Raw Simulation (RawSim): Raw simulator output; error reflects influence of domain characteristics $\mathbf{c}$ mismatch.
    \item Residual Dynamics (ResDyn): Predicts residual corrections using observation history for forward inference.
    \item Recurrent State Space Model (RSSM): An explicit recurrent latent inference model that maintains a learned latent state from observation history, representing a more expressive history-based baseline.
\end{itemize}

\subsubsection{Evaluation Tasks}
\begin{figure}[t]
    \centering
    \includegraphics[width=\linewidth]{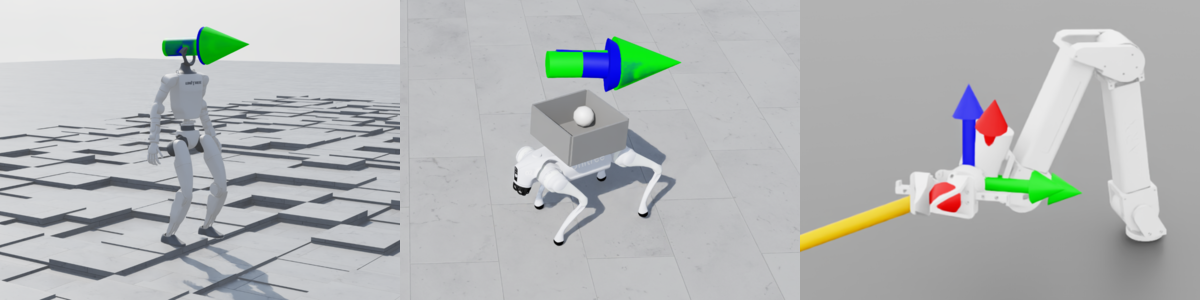}
    \caption{Simulation environments. Left: G1 humanoid on irregular terrain. Middle: Go2 quadruped with payload (ball in box). Right: R5 manipulator under external forces.}
    \label{fig:sim-demo}
    \vspace{-15pt}
\end{figure}

We design custom tasks on three robotic platforms (Figure~\ref{fig:sim-demo}), each with two conditions: one where $\mathbf{h}_t$ has minimal influence (low-$\mathbf{h}_t$) and one where $\mathbf{h}_t$ significantly affects transitions (high-$\mathbf{h}_t$).
All platforms use proprioceptive observations. Contact forces, terrain geometry, and payload motion are therefore not observable in our setting.

\begin{itemize}
    \item G1 Humanoid: Velocity tracking on flat terrain (low-$\mathbf{h}_t$) vs.\ irregular stepping blocks with unpredictable foot contacts (high-$\mathbf{h}_t$).
    \item \label{sec:go2-task}Go2 Quadruped: Locomotion without payload (low-$\mathbf{h}_t$) vs.\ with 5kg ball in mounted box creating unpredictable interaction forces (high-$\mathbf{h}_t$).
    \item \label{sec:r5-task}R5 Manipulator: End-effector pose tracking in its workspace (low-$\mathbf{h}_t$) vs.\ under 0--200N external force disturbances (high-$\mathbf{h}_t$).
\end{itemize}

\subsubsection{Single-Step Rollout}

\begin{figure*}[t]
    \centering
\begin{tikzpicture}[>=Stealth]
    \begin{scope}[
            shift={(-6.0cm, -0.2cm)},
            xscale=0.7,
            yscale=0.85,
            every node/.style={font=\small},
            thick
        ]
        \node (o1) at (0, -0.4) {$o^R_1,$};
        \node[anchor=west, inner sep=1pt] (a1) at (o1.east) {$a_1$};
        \node (o2) at (3.0, -0.4) {$o^R_2,$};
        \node[anchor=west, inner sep=1pt] (a2) at (o2.east) {$a_2$};
        \node (o3) at (6.0, -0.4) {$o^R_3$};

        \node (o2s) at ([yshift=3.2cm]o2.center) {$o^S_2$};
        \node (o2r) at ([yshift=1.6cm]o2.center) {$o^{S \to R}_2$};

        \node (o3s) at ([yshift=3.2cm]o3.center) {$o^S_3$};
        \node (o3r) at ([yshift=1.6cm]o3.center) {$o^{S \to R}_3$};

        \draw[->, black] (a1) -- (o2.west);
        \draw[->, black] (a1) -- (o2s);
        \draw[->, blue] (o2s) -- (o2r);
        \draw[->, blue] (a1) -- (o2r);
        \draw[<->, red] (o2r) -- node[right] {$e$} (o2);

        \draw[->, black] (a2) -- (o3.west);
        \draw[->, black] (a2) -- (o3s);
        \draw[->, blue] (o3s) -- (o3r);
        \draw[->, blue] (a2) -- (o3r);
        \draw[<->, red] (o3r) -- node[right] {$e$} (o3);

        \node[anchor=north west, font=\scriptsize] at (-0.7, 3.0) {
            \begin{tabular}{@{}l@{}}
                \tikz[baseline=-0.5ex]{\draw[->, blue, thick] (0,0) -- (0.5,0);}  \\ Model inference \\[2pt]
                \tikz[baseline=-0.5ex]{\draw[->, black, thick] (0,0) -- (0.5,0);} \\ System step
            \end{tabular}
        };
    \end{scope}

    \begin{axis}[
            name=g1,
            ybar,
            bar width=4pt,
            width=0.26\textwidth,
            height=3.5cm,
            ylabel={MSE},
            ylabel style={font=\scriptsize},
            title={G1 Humanoid},
            title style={font=\small},
            symbolic x coords={Flat, Uneven},
            xtick=data,
            xticklabel style={font=\scriptsize},
            ymin=0,
            yticklabel style={font=\tiny, /pgf/number format/fixed},
            axis lines=left,
            axis line style={-},
            enlarge x limits=0.6,
            every axis plot/.append style={fill opacity=0.8},
            clip=false,
        ]
        \addplot[fill=gray!60, draw=gray!80] coordinates {(Flat, 0.030434) (Uneven, 0.052726)};
        \addplot[fill=gray!100, gray!100] coordinates {(Flat, 0.019182) (Uneven, 0.119039)};
        \addplot[fill=blue!40, draw=blue!70] coordinates {(Flat, 0.006147) (Uneven, 0.070)};  
        \addplot[fill=orange!60, draw=orange!80] coordinates {(Flat, 0.020385) (Uneven, 0.055421)};
        \addplot[fill=red!60, draw=red!80] coordinates {(Flat, 0.007184) (Uneven, 0.034837)};
    \end{axis}

    \begin{axis}[
            at={(g1.east)},
            anchor=west,
            xshift=0.8cm,
            name=go2,
            ybar,
            bar width=4pt,
            width=0.26\textwidth,
            height=3.5cm,
            ylabel={MSE},
            ylabel style={font=\scriptsize},
            title={Go2 Quadruped},
            title style={font=\small},
            symbolic x coords={Flat, Payload},
            xtick=data,
            xticklabel style={font=\scriptsize},
            ymin=0,
            yticklabel style={font=\tiny, /pgf/number format/fixed},
            axis lines=left,
            axis line style={-},
            enlarge x limits=0.6,
            every axis plot/.append style={fill opacity=0.8},
            clip=false,
        ]
        \addplot[fill=gray!60, draw=gray!80] coordinates {(Flat, 0.365171) (Payload, 0.305472)};
        \addplot[fill=gray!100, gray!100] coordinates {(Flat, 0.349060) (Payload, 0.574078)};
        \addplot[fill=blue!40, draw=blue!70] coordinates {(Flat, 0.414751) (Payload, 0.298)};  
        \addplot[fill=orange!60, draw=orange!80] coordinates {(Flat, 0.441840) (Payload, 0.492323)};
        \addplot[fill=red!60, draw=red!80] coordinates {(Flat, 0.126424) (Payload, 0.147927)};
    \end{axis}

    \begin{axis}[
            at={(go2.east)},
            anchor=west,
            xshift=0.8cm,
            ybar,
            bar width=4pt,
            width=0.26\textwidth,
            height=3.5cm,
            ylabel={MSE},
            ylabel style={font=\scriptsize},
            title={R5 Manipulator},
            title style={font=\small},
            symbolic x coords={Normal, Force},
            xtick=data,
            xticklabel style={font=\scriptsize},
            ymin=0,
            yticklabel style={font=\tiny, /pgf/number format/fixed},
            axis lines=left,
            axis line style={-},
            enlarge x limits=0.6,
            every axis plot/.append style={fill opacity=0.8},
            clip=false,
            legend style={
                    at={(-1.0,-0.25)},
                    anchor=north,
                    legend columns=5,
                    font=\scriptsize,
                    draw=none,
                    column sep=8pt,
                },
        ]
        \addplot[fill=gray!60, draw=gray!80] coordinates {(Normal, 0.238427) (Force, 0.706230)};
        \addplot[fill=gray!100, gray!100] coordinates {(Normal, 0.380090) (Force, 0.741280)};
        \addplot[fill=blue!40, draw=blue!70] coordinates {(Normal, 0.185424) (Force, 0.515493)};  
        \addplot[fill=orange!60, draw=orange!80] coordinates {(Normal, 0.134410) (Force, 0.762682)};
        \addplot[fill=red!60, draw=red!80] coordinates {(Normal, 0.163895) (Force, 0.401467)};

        \legend{RawSim, DirectPred, RSSM, ResDyn, Ours}
    \end{axis}
\end{tikzpicture}
    \caption{Dynamics modeling evaluation. Left: teacher-forcing protocol where the model predicts translated observations $o^{S \to R}$ from simulator transitions, with error $e$ measured against ground-truth. Right: prediction error (MSE) across three platforms. For each robot, the left bar group shows the low-$\mathbf{h}_t$ task, and the right shows the high-$\mathbf{h}_t$ task. DirectPred and RawSim characterize the two challenges: hidden variable influence and domain gap; RSSM and ResDyn are stronger history-based and residual baselines. World Translation consistently improves over all baselines, with the largest gains where hidden variable influence is strongest.}
    \label{fig:modeling-error}
    \vspace{-10pt}
\end{figure*}

We measure single-step prediction error under teacher-forcing, where the model receives ground-truth observations at each step. This isolates dynamics modeling accuracy, since both alternatives for autoregressive evaluation introduce confounds: ground-truth actions are computed for a different state, and recomputing actions per method couples policy behavior with dynamics error.

Figure~\ref{fig:modeling-error} shows the dynamics modeling accuracy across platforms. DirectPred and RawSim reveal the structure of each task: Direct Prediction error reflects hidden variable influence, while Raw Simulation error reflects domain gap severity.
World Translation consistently outperforms both references across all conditions. It recovers hidden variable information (improving over Direct Prediction) while correcting domain mismatch (improving over Raw Simulation). In high-$\mathbf{h}_t$ conditions where DirectPred error is substantial, World Translation achieves the largest improvements (G1 Uneven: 0.035 vs.\ DirectPred 0.119; R5 Force: 0.401 vs.\ DirectPred 0.741).
Residual Dynamics shows mixed results: it improves over RawSim in some conditions but shows limited gains over DirectPred in high-$\mathbf{h}_t$ tasks. RSSM, as a more expressive history-based model, generally reduces error compared to DirectPred, with the most notable gains in high-$\mathbf{h}_t$ conditions. Yet in high-$\mathbf{h}_t$ conditions it remains substantially above World Translation (Go2 Payload: 0.298 vs.\ 0.148; R5 Force: 0.515 vs.\ 0.401), showing that the gap reflects a limitation of history-based inference itself rather than model capacity. Backward extraction addresses the underlying issue by inferring from outcomes.

\subsubsection{Multi-Step Rollout}

Teacher-forcing isolates dynamics accuracy but does not capture error compounding under closed-loop deployment. We therefore evaluate autoregressive rollouts on the Go2 high-$\mathbf{h}_t$ task, recomputing policy actions on predicted states at each step. Unlike the single-step evaluation, policy-dynamics coupling is expected here, as we are assessing closed-loop stability rather than isolated prediction quality.

\begin{figure}[t]
    \centering
    \includegraphics[width=\linewidth]{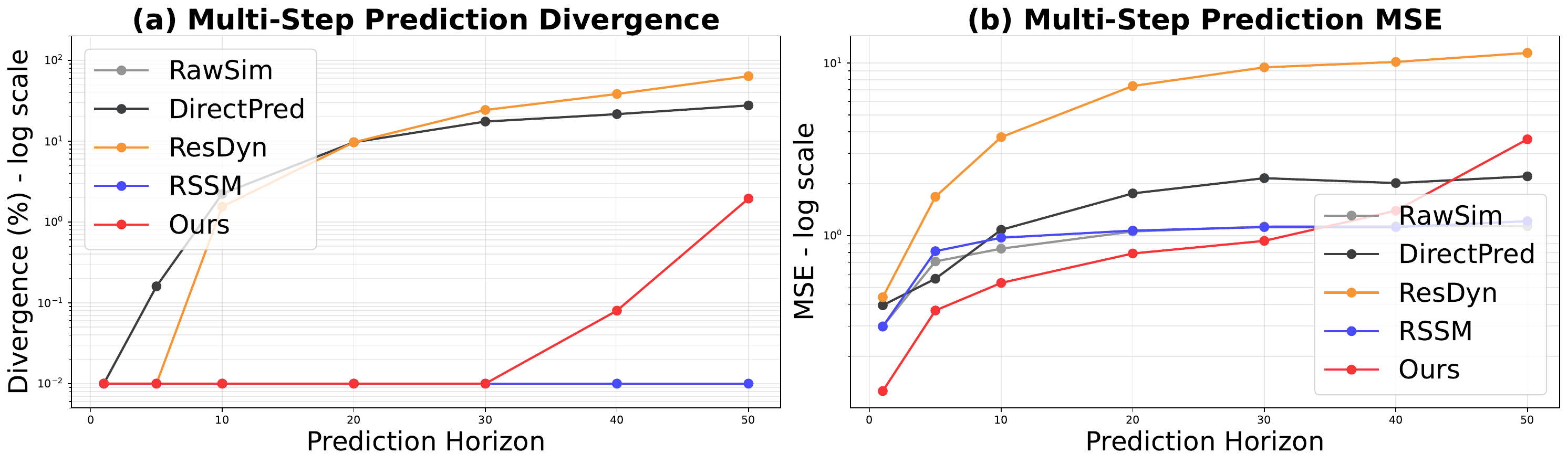}
    \vspace{-10pt}
    \caption{Multi-step rollout on Go2 high-$\mathbf{h}_t$ task. (a) Divergence rate (fraction of rollouts where base height or velocity exceeds physical bounds). (b) Prediction MSE. RSSM and RawSim show zero divergence but RSSM MSE remains above RawSim; World Translation remains stable through horizon 30 with only $\sim$2\% divergence at horizon 50.}
    \label{fig:multistep}
    \vspace{-10pt}
\end{figure}

Figure~\ref{fig:multistep} shows divergence rate and MSE over rollout horizons. RawSim shows zero divergence by construction and its MSE saturates as both predicted and ground-truth trajectories stay bounded; exceeding this level indicates non-physical rollout behavior. Without $\mathbf{h}_t$ inference, ResDyn and DirectPred diverge rapidly; ResDyn in particular can overwrite physically invalid contact states, leading to out-of-distribution inputs. RSSM shows zero divergence but its MSE remains slightly above RawSim, without improving over the raw physics baseline. World Translation remains stable through horizon 30 and exhibits only $\sim$2\% divergence at horizon 50.

For MSE (Figure~\ref{fig:multistep}b), World Translation achieves the lowest error up to horizon $\sim$30. At longer horizons, state-overwriting methods accumulate numerical errors as corrected states feed back into the physics engine, raising MSE above DirectPred. However, DirectPred's lower long-horizon MSE must be interpreted alongside its $\sim$20\% divergence rate; accurate predictions on diverged rollouts are not meaningful. World Translation provides the best accuracy--stability tradeoff at operationally relevant horizons.

\subsection{Representation Analysis}
\label{sec:representation-analysis}

We analyze the representation on the ARX R5 manipulator (Section~\ref{sec:r5-task}), where external forces ($\mathbf{h}_t$) and motor parameters ($\mathbf{c}$) interact: the same force produces different joint responses depending on actuator stiffness.

\subsubsection{Hidden Variable Information $\mathbf{h}_t$}
\label{sec:info-exp}

To verify that transitions preserve hidden variable information, we train lightweight decoders to predict the ground-truth hidden variable $\mathbf{h}_t$ (normalized external force) from different input representations. Table~\ref{tab:info-preservation} reports the $R^2$ score for predicting force magnitude.
\begin{table}[t]
    \centering
    \caption{Hidden variable ($\mathbf{h}_t$) prediction $R^2$ from different representations. Combined $(\mathbf{o}_t, \mathbf{a}_t, \mathbf{z}_t)$ results compare w/o (basic VAE) vs Full model, showing that auxiliary components do not compromise $\mathbf{h}_t$ preservation.}
    \scriptsize
    \label{tab:info-preservation}
    \begin{tabular}{lcc}
        \hline
        Input                                                    & w/o                      & Full \\
        \hline
        \multicolumn{3}{l}{\textit{Baseline}}                                                      \\
        $(\mathbf{o}, \mathbf{a})^S$                             & \multicolumn{2}{c}{0.73}        \\
        $(\mathbf{o}, \mathbf{a})^R$                             & \multicolumn{2}{c}{0.66}        \\
        $(\mathbf{o}_t, \mathbf{a}_t, \mathbf{o}_{t+1})^S$       & \multicolumn{2}{c}{0.78}        \\
        $(\mathbf{o}_t, \mathbf{a}_t, \mathbf{o}_{t+1})^R$       & \multicolumn{2}{c}{0.73}        \\
        \hline
        \multicolumn{3}{l}{\textit{Combined $(\mathbf{o}_t, \mathbf{a}_t, \mathbf{z}_t)$}}         \\
        $(\mathbf{o}, \mathbf{a})^S, \mathbf{z}^S$               & 0.81                     & 0.80 \\
        $(\mathbf{o}, \mathbf{a})^R, \mathbf{z}^R$               & 0.68                     & 0.72 \\
        $(\mathbf{o}, \mathbf{a})^S, \mathbf{z}^{S \to R}$       & 0.67                     & 0.68 \\
        $(\mathbf{o}, \mathbf{a})^R, \mathbf{z}^{R \to S}$       & 0.75                     & 0.78 \\
        $(\mathbf{o}, \mathbf{a})^S, \mathbf{z}^{S \to R \to S}$ & 0.78                     & 0.78 \\
        $(\mathbf{o}, \mathbf{a})^R, \mathbf{z}^{R \to S \to R}$ & 0.67                     & 0.67 \\
        \hline
    \end{tabular}
    \vspace{-15pt}
\end{table}
$(\mathbf{o}_t, \mathbf{a}_t)$ alone explains 66--73\% of $\mathbf{h}_t$ variance (confirming leakage), while $(\mathbf{o}_t, \mathbf{a}_t, \mathbf{z}_t)$ matches the full transition's predictive power (68--81\% vs.\ 73--78\%). After translation, predictive power is largely preserved (67--78\%), and comparing w/o vs.\ Full columns shows the auxiliary components add $\mathbf{c}$ to $\mathbf{z}_t$ without compromising $\mathbf{h}_t$ retention.

\subsubsection{Domain Characteristic Information $\mathbf{c}$}
\label{sec:domain-char-exp}

To verify that our auxiliary components successfully encourage $\mathbf{z}_t$ to encode domain characteristics, we conduct domain classification experiments. We train a binary classifier as a probe to measure how much domain information each representation contains. The classifier is trained with ground-truth domain labels and directly predicts S or R from different input representations. We evaluate on: (1) $(\mathbf{o}_t, \mathbf{a}_t)$ alone to measure baseline information leakage, (2) $\mathbf{z}_t$ alone to measure how much domain information the encoder captures, and (3) combined $(\mathbf{o}_t, \mathbf{a}_t, \mathbf{z}_t)$ to measure whether $\mathbf{z}_t$ adds domain information beyond $(\mathbf{o}_t, \mathbf{a}_t)$. For the combined input, $(\mathbf{o}_t, \mathbf{a}_t)$ is always from the source domain to match the practical application scenario where we observe transitions in one domain and translate $\mathbf{z}_t$ to another.
\begin{table}[t]
    \centering
    \caption{Domain classification accuracy from different representations. ``w/o'' denotes the basic VAE without domain classifier and FiLM; ``Full'' denotes the complete model.}
    \label{tab:domain-char}
    \scriptsize
    \begin{tabular}{llcc}
        \hline
        Input                                                    & Target & w/o                        & Full   \\
        \hline
        \multicolumn{4}{l}{\textit{Information in $(\mathbf{o}_t, \mathbf{a}_t)$}}                              \\
        $(\mathbf{o}, \mathbf{a})^S$                             & S      & \multicolumn{2}{c}{0.7894}          \\
        $(\mathbf{o}, \mathbf{a})^R$                             & R      & \multicolumn{2}{c}{0.8479}          \\
        \hline
        \multicolumn{4}{l}{\textit{Information in $\mathbf{z}_t$}}                                              \\
        $\mathbf{z}^S$                                           & S      & 0.6729                     & 0.9407 \\
        $\mathbf{z}^R$                                           & R      & 0.6412                     & 0.9365 \\
        $\mathbf{z}^{S \to R}$                                   & R      & 0.4288                     & 0.9124 \\
        $\mathbf{z}^{R \to S}$                                   & S      & 0.4936                     & 0.8327 \\
        $\mathbf{z}^{S \to R \to S}$                             & S      & 0.7015                     & 0.8427 \\
        $\mathbf{z}^{R \to S \to R}$                             & R      & 0.6099                     & 0.9820 \\
        \hline
        \multicolumn{4}{l}{\textit{Combined $(\mathbf{o}_t, \mathbf{a}_t, \mathbf{z}_t)$}}                      \\
        $(\mathbf{o}, \mathbf{a})^S, \mathbf{z}^S$               & S      & 0.7857                     & 0.9443 \\
        $(\mathbf{o}, \mathbf{a})^R, \mathbf{z}^R$               & R      & 0.8326                     & 0.9414 \\
        $(\mathbf{o}, \mathbf{a})^S, \mathbf{z}^{S \to R}$       & R      & 0.2702                     & 0.9070 \\
        $(\mathbf{o}, \mathbf{a})^R, \mathbf{z}^{R \to S}$       & S      & 0.2404                     & 0.8309 \\
        $(\mathbf{o}, \mathbf{a})^S, \mathbf{z}^{S \to R \to S}$ & S      & 0.7952                     & 0.8459 \\
        $(\mathbf{o}, \mathbf{a})^R, \mathbf{z}^{R \to S \to R}$ & R      & 0.8227                     & 0.9837 \\
        \hline
    \end{tabular}
\end{table}
Table~\ref{tab:domain-char} shows that without auxiliary components, $\mathbf{z}_t$ achieves only 64--67\% domain classification (worse than $(\mathbf{o}_t, \mathbf{a}_t)$ alone at 79--85\%), and translated $\mathbf{z}^{S \to R}$ classifies at only 43\% as R (below chance). With auxiliary components, $\mathbf{z}_t$ reaches 94\% and $\mathbf{z}^{S \to R}$ reaches 91\% as R, showing that the translated feature successfully overrides the domain signal in $(\mathbf{o}_t, \mathbf{a}_t)$. Auxiliary components are essential: without them, cross-domain prediction fails.

\begin{figure}[t]
    \centering
    \includegraphics[width=0.9\linewidth]{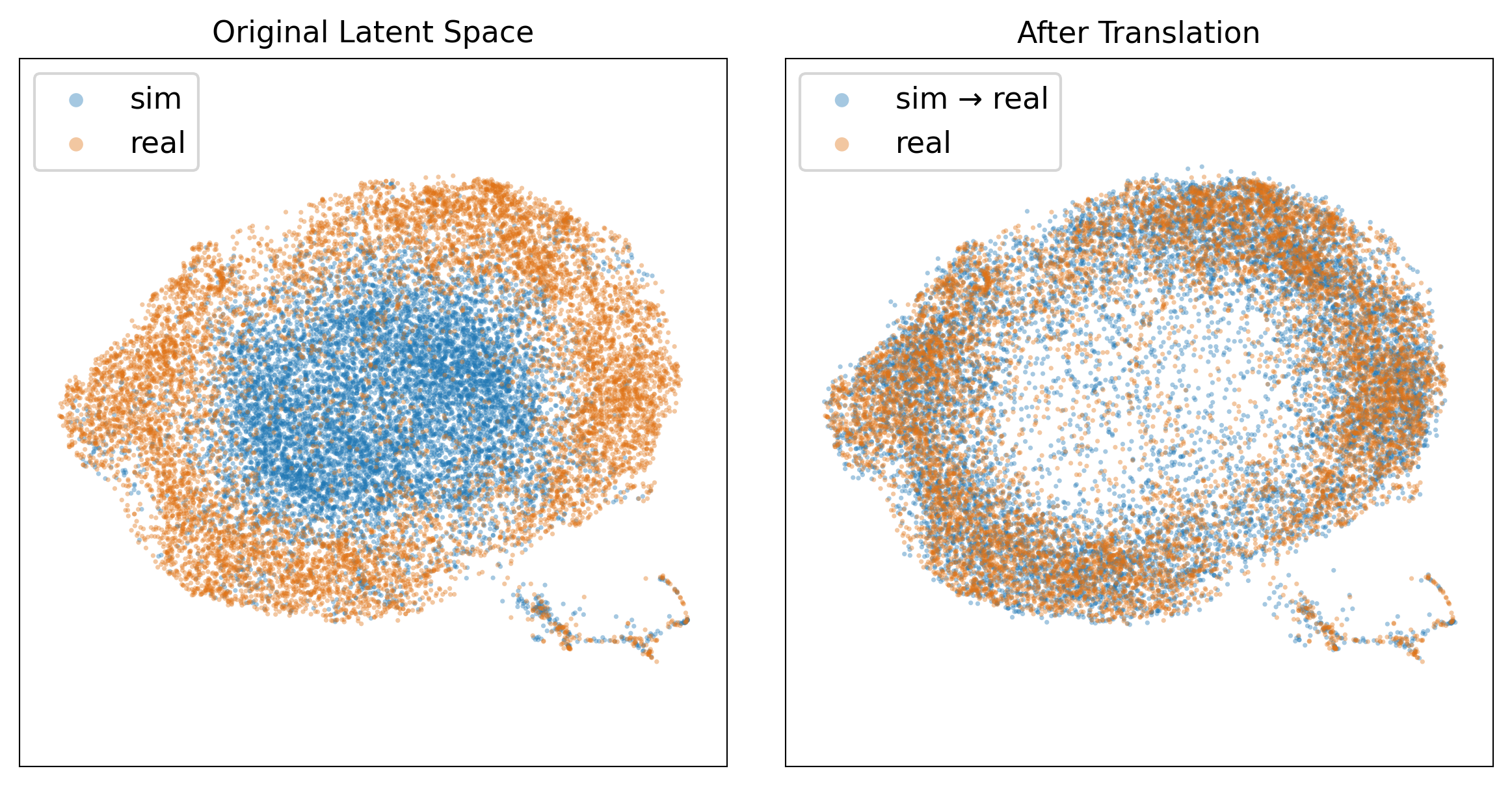}
    \caption{UMAP visualization of dynamics features on the R5 manipulator task. Left: original latent space shows domain separation reflecting $\mathbf{c}$ encoded in $\mathbf{z}_t$. Right: after translation, $\mathbf{z}^{S \to R}$ (blue) aligns with real domain $\mathbf{z}^R$ (orange).}
    \label{fig:latent-space}
    \vspace{-15pt}
\end{figure}

Figure~\ref{fig:latent-space} visualizes the dynamics features using UMAP~\cite{mcinnes2018umap}: the original latent space shows clear domain separation (required for translation to learn meaningful mappings), and after translation, $\mathbf{z}^{S \to R}$ aligns with real domain features $\mathbf{z}^R$. Comparing w/o and Full columns in Table~\ref{tab:info-preservation} confirms the auxiliary components \textit{add} domain information $\mathbf{c}$ without removing $\mathbf{h}_t$.

\subsection{Ablation Study}
\label{sec:ablation}

We study the necessity of key design choices in World Translation. Experiments are conducted on the ARX R5 manipulator unless otherwise noted.

\subsubsection{Translation in Latent Space}

\begin{figure}[t]
    \centering
    \includegraphics[width=0.95\linewidth]{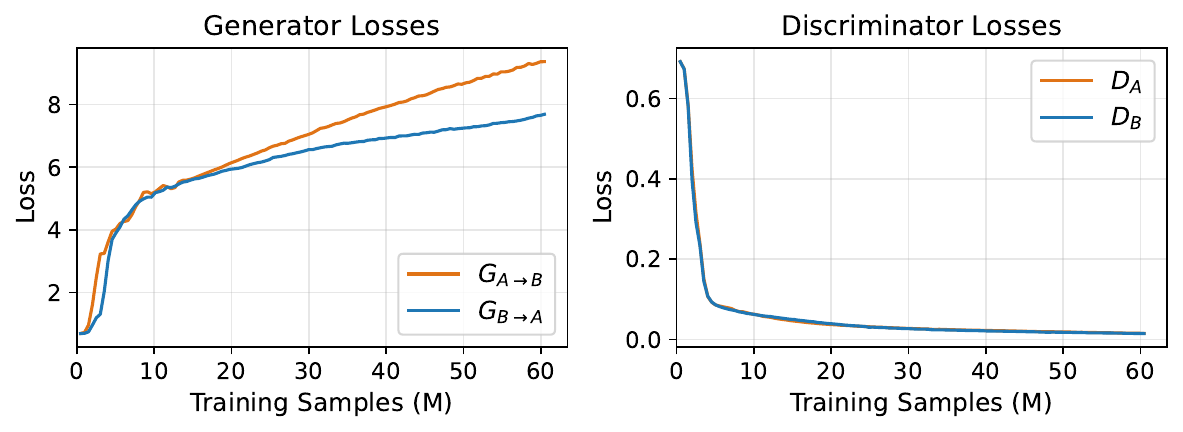}
    \caption{Training collapse in observation-level translation. Discriminator losses (right) drop to near-zero while generator losses (left) diverge, indicating the discriminators trivially distinguish domains using $(\mathbf{o}_t, \mathbf{a}_t)$ without learning meaningful dynamics mappings.}
    \label{fig:translation-collapse}
\end{figure}

We justify translating $\mathbf{z}_t$ rather than observations directly by training a generator that maps $(\mathbf{o}_t, \mathbf{a}_t, \mathbf{o}_{t+1}^A) \to \mathbf{o}_{t+1}^B$ at the observation level. This collapses immediately (Figure~\ref{fig:translation-collapse}): since $(\mathbf{o}_t, \mathbf{a}_t)$ alone achieves 79--85\% domain classification, the discriminator trivially distinguishes domains without examining $\mathbf{o}_{t+1}$, providing no useful gradient. Abstracting to $\mathbf{z}_t$ isolates dynamics-relevant information from the domain-leaking $(\mathbf{o}_t, \mathbf{a}_t)$, enabling stable training. Furthermore, observation-level translation has no clear objective: without our abstract representation, there is no criterion for which dynamics properties a translated $\mathbf{o}_{t+1}$ should preserve.

\subsubsection{Translation Components}
We ablate components of the CycleGAN translation network; results are in Table~\ref{tab:ablation-translation}.
\begin{table}[t]
    \centering
    \caption{Ablation of translation components. Cycle Error measures round-trip reconstruction $\|\mathbf{z} - G(G(\mathbf{z}))\|$; Identity Error measures unnecessary modification $\|\mathbf{z}^R - G_{S \to R}(\mathbf{z}^R)\|$.}
    \label{tab:ablation-translation}
    \begin{tabular}{lccc}
        \hline
        Variant               & Cycle Error & Identity Error & Pred. Error \\
        \hline
        Full Model            & 0.034       & 0.233          & 0.401       \\
        w/o Cycle Consistency & 0.273       & 0.004          & 0.681       \\
        w/o Identity Loss     & 0.026       & 0.463          & 0.483       \\
        \hline
    \end{tabular}
    \vspace{-10pt}
\end{table}
Without cycle consistency, cycle error increases 8$\times$ (0.034 $\to$ 0.273) and prediction error degrades significantly; identity error drops to near-zero, indicating the generator becomes overly conservative. Without identity loss, identity error doubles (0.233 $\to$ 0.463), showing unnecessary distortion of in-domain samples.

\subsection{Downstream Policy Performance on Real Robot}
\label{sec:downstream}

\begin{figure}[t]
    \centering
    \includegraphics[width=\linewidth]{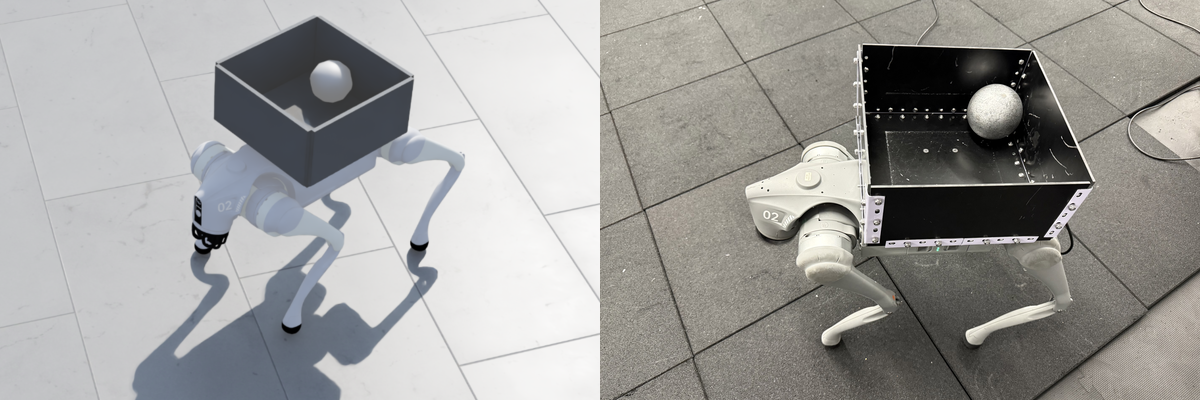}
    \caption{Real-world deployment of the Go2 quadruped with the payload task. Left: simulation setup for training, right: real-world deployment for evaluation.}
    \label{fig:real-demo}
    \vspace{-15pt}
\end{figure}

We conducted a real robot experiment to further evaluate whether improved dynamics modeling translates to better policy transfer in a practical sim-to-real deployment scenario. The real-world experiment setup is illustrated in Figure~\ref{fig:real-demo}.

\subsubsection{Setup}

We evaluate on the real Go2 quadruped with the payload task described earlier in Section~\ref{sec:go2-task}, where a ball moves freely inside a box mounted on the robot. All methods start from a base policy trained in simulation without domain randomization. This base policy is then fine-tuned using different adaptation strategies. The baselines are:
\begin{itemize}
    \item No Adaptation: The base policy is deployed directly without fine-tuning.
    \item Domain Randomization: Fine-tuned with randomized physics parameters (base mass $-$1 to $+$3 kg, friction 0.3--1.2, actuator gains 0.8--1.25$\times$) and additive observation noise for 1000 steps.
    \item World Translation (Ours): Fine-tuned using translated dynamics. We collect 20 minutes of real-world transitions using a joystick-controlled base policy, train the translation model, then fine-tune the policy for 1000 gradient update iterations using translated dynamics predictions. Unpaired training reduces data requirements since transitions need only cover the relevant dynamics regime, unlike paired approaches that require matching samples across domains.
\end{itemize}

\subsubsection{Metrics}
All policies track the same recorded commanded angular velocity sequence, enabling controlled comparison. We evaluate using metrics computable from proprioceptive observations, each corresponding to a reward term during training:
\begin{itemize}
    \item Angular Velocity Tracking Error: RMSE between commanded and measured angular velocity. Lower is better.
    \item Orientation Stability: Root mean square of pitch and roll standard deviations, $\sqrt{\sigma_\text{pitch}^2 + \sigma_\text{roll}^2}$. Lower is better.
    \item Average Power: Mean joint power consumption $\frac{1}{T}\int |\boldsymbol{\tau} \cdot \dot{\mathbf{q}}| \, dt$. Lower is better.
\end{itemize}

\subsubsection{Results}

\begin{table}[t]
    \centering
    \caption{Downstream policy performance on Go2 with payload. All methods track the same angular velocity command.}
    \label{tab:downstream}
    \begin{tabular}{lccc}
        \hline
        Method            & Tracking (rad/s) & Stability (rad) & Power (W) \\
        \hline
        No Adaptation     & 0.2475           & 0.0517          & 28.07     \\
        Domain Rand.      & 0.3306           & 0.0694          & 21.95     \\
        World Translation & 0.1786           & 0.0513          & 29.55     \\
        \hline
    \end{tabular}
    \vspace{-10pt}
\end{table}

Table~\ref{tab:downstream} shows World Translation achieves the lowest tracking error (0.18 vs.\ 0.25 No Adaptation, 0.33 Domain Randomization) while maintaining comparable stability and power. Domain Randomization's lower power consumption reflects overly conservative behavior that sacrifices responsiveness. These results confirm that improved dynamics modeling transfers to better command tracking, though effects on other metrics are modest.

\section{Discussion}
\label{sec:discussion}

\subsection{Assumptions and Scope}

Backward dynamics extraction assumes that all dynamics information not captured in $(\mathbf{o}_t, \mathbf{a}_t)$ is recoverable from the full transition $(\mathbf{o}_t, \mathbf{a}_t, \mathbf{o}_{t+1})$. This holds when observations are high-dimensional relative to the hidden variable effects and when $\mathbf{h}_t$ creates distinct signatures in the outcome. Discrete contact events satisfy both conditions naturally; continuous forces have subtler effects and leave weaker signatures. Table~\ref{tab:info-preservation} provides empirical support: the full transition achieves $R^2$ of 0.73--0.78 for predicting external forces on the R5 task, compared to 0.66--0.73 from $(\mathbf{o}_t, \mathbf{a}_t)$ alone.

A harder constraint is \textit{unmodeled phenomena}: if a physical effect is entirely absent from the simulator, no translation can recover it. The simulator must capture the relevant physics, even approximately. Translation corrects \textit{how} these effects manifest in dynamics, not whether they exist.

\subsection{Design Tradeoffs}

\textit{Unpaired learning versus direct supervision.}
World Translation optimizes reconstruction, adversarial, and cycle-consistency losses on unpaired data, rather than minimizing prediction error on paired samples. This indirect optimization may sacrifice accuracy relative to direct supervision, but paired sim-real transitions with matching hidden variables are generally impossible to obtain. The tradeoff enables learning from data that would otherwise be unusable.

\textit{Implicit representation.}
The learned $\mathbf{z}_t$ captures effects of $\mathbf{h}_t$ and $\mathbf{c}$ without explicit parameterization, allowing it to encode unmodeled dynamics that system identification cannot reach. The cost is interpretability: diagnosing training failures or verifying translation quality requires indirect proxies (domain classification accuracy, downstream prediction error) rather than direct inspection of the representation.

These tradeoffs interact in practice. Adversarial training is less stable than supervised learning, and the absence of intuitive evaluation criteria makes it harder to detect silent failure modes in the latent space.

\subsection{Practical Challenges}

\textit{Tight coupling between $\mathbf{h}_t$ and $\mathbf{c}$.}
Translation is harder when hidden variables and domain characteristics interact. On the R5 task, external forces couple with domain-specific motor gains: the same force produces different joint responses depending on actuator stiffness, making it difficult to separate what should be translated from what should be preserved. This explains the smaller improvement margins on R5 compared to locomotion tasks, where contact events are more naturally separable from motor parameters.

\textit{State overwriting.}
Our alignment procedure overwrites observable states while leaving hidden simulator buffers unchanged, which can cause inconsistent physics decisions when internal state depends on observable state history. Multi-step results (Section~\ref{sec:experiments}) show the effect remains bounded at operationally relevant horizons. Action correction methods~\cite{hanna2017grounded, karnan2020rgat, he2025asap} avoid this by modifying actions rather than states; combining action correction with World Translation's translated targets is a natural extension that would keep internal simulator state consistent.

\textit{Training stability.}
Joint training of a VAE, blind decoder, domain classifier, and CycleGAN involves competing objectives that require careful balancing. Empirically, the same hyperparameters transfer across all three platforms without per-platform tuning (Appendix Table II), suggesting reasonable stability in practice.

\subsection{Future Directions}

\textit{Multi-step training.} Our single-step formulation accumulates prediction errors over long horizons. Training on trajectory segments with autoregressive objectives could reduce compounding errors.

\textit{Online adaptation.} Updating the translation model during deployment would address distribution shift without offline retraining. The unpaired formulation is well-suited to this: new real-world transitions can be incorporated without requiring simulated counterparts.

\textit{Visual observations.} Extending to image-based observations would support translation for vision-based policies, requiring disentanglement of visual appearance from dynamics-relevant content.

\section{Conclusion}

We presented World Translation, a framework for learned dynamics models that addresses sim-to-real transfer by reformulating dynamics alignment as unpaired domain translation over learned features. Backward dynamics extraction captures hidden variable effects from observed transitions without assuming that history is sufficient to recover them; cycle-consistent translation then maps these features across domains while preserving dynamics information. Experiments on humanoid, quadruped, and manipulator platforms demonstrate improved dynamics modeling, with the largest gains when history is insufficient to recover hidden variable effects, and real-robot deployment on Go2 shows improved command tracking.

\bibliographystyle{IEEEtran}
\bibliography{references}

@article{lee2020learning,
  title     = {Learning quadrupedal locomotion over challenging terrain},
  author    = {Lee, Joonho and Hwangbo, Jemin and Wellhausen, Lorenz and Koltun, Vladlen and Hutter, Marco},
  journal   = {Science robotics},
  volume    = {5},
  number    = {47},
  pages     = {eabc5986},
  year      = {2020},
  publisher = {American Association for the Advancement of Science}
}

@article{hwangbo2019learning,
  title     = {Learning agile and dynamic motor skills for legged robots},
  author    = {Hwangbo, Jemin and Lee, Joonho and Dosovitskiy, Alexey and Bellicoso, Dario and Tsounis, Vassilios and Koltun, Vladlen and Hutter, Marco},
  journal   = {Science Robotics},
  volume    = {4},
  number    = {26},
  pages     = {eaau5872},
  year      = {2019},
  publisher = {American Association for the Advancement of Science}
}

@article{singh2024dextrah,
  title   = {Dextrah-rgb: Visuomotor policies to grasp anything with dexterous hands},
  author  = {Singh, Ritvik and Allshire, Arthur and Handa, Ankur and Ratliff, Nathan and Van Wyk, Karl},
  journal = {arXiv preprint arXiv:2412.01791},
  year    = {2024}
}

@inproceedings{tobin2017domain,
  title        = {Domain randomization for transferring deep neural networks from simulation to the real world},
  author       = {Tobin, Josh and Fong, Rachel and Ray, Alex and Schneider, Jonas and Zaremba, Wojciech and Abbeel, Pieter},
  booktitle    = {2017 IEEE/RSJ international conference on intelligent robots and systems (IROS)},
  pages        = {23--30},
  year         = {2017},
  organization = {IEEE}
}

@inproceedings{peng2018sim,
  title        = {Sim-to-real transfer of robotic control with dynamics randomization},
  author       = {Peng, Xue Bin and Andrychowicz, Marcin and Zaremba, Wojciech and Abbeel, Pieter},
  booktitle    = {2018 IEEE international conference on robotics and automation (ICRA)},
  pages        = {3803--3810},
  year         = {2018},
  organization = {IEEE}
}

@article{akkaya2019solving,
  title   = {Solving rubik's cube with a robot hand},
  author  = {Akkaya, Ilge and Andrychowicz, Marcin and Chociej, Maciek and Litwin, Mateusz and McGrew, Bob and Petron, Arthur and Paino, Alex and Plappert, Matthias and Powell, Glenn and Ribas, Raphael and others},
  journal = {arXiv preprint arXiv:1910.07113},
  year    = {2019}
}

@inproceedings{josifovski2022analysis,
  title        = {Analysis of randomization effects on sim2real transfer in reinforcement learning for robotic manipulation tasks},
  author       = {Josifovski, Josip and Malmir, Mohammadhossein and Klarmann, Noah and {\v{Z}}agar, Bare Luka and Navarro-Guerrero, Nicol{\'a}s and Knoll, Alois},
  booktitle    = {2022 IEEE/RSJ International Conference on Intelligent Robots and Systems (IROS)},
  pages        = {10193--10200},
  year         = {2022},
  organization = {IEEE}
}

@article{tiboni2023dropo,
  title     = {DROPO: Sim-to-real transfer with offline domain randomization},
  author    = {Tiboni, Gabriele and Arndt, Karol and Kyrki, Ville},
  journal   = {Robotics and Autonomous Systems},
  volume    = {166},
  pages     = {104432},
  year      = {2023},
  publisher = {Elsevier}
}

@inproceedings{mehta2020active,
  title        = {Active domain randomization},
  author       = {Mehta, Bhairav and Diaz, Manfred and Golemo, Florian and Pal, Christopher J and Paull, Liam},
  booktitle    = {Conference on Robot Learning},
  pages        = {1162--1176},
  year         = {2020},
  organization = {PMLR}
}

@article{xiao2025learning,
  title   = {Learning robotic policy with imagined transition: Mitigating the trade-off between robustness and optimality},
  author  = {Xiao, Wei and Lyu, Shangke and Gong, Zhefei and Wang, Renjie and Wang, Donglin},
  journal = {arXiv preprint arXiv:2503.10484},
  year    = {2025}
}

@incollection{ljung1998system,
  title     = {System identification},
  author    = {Ljung, Lennart},
  booktitle = {Signal analysis and prediction},
  pages     = {163--173},
  year      = {1998},
  publisher = {Springer}
}

@inproceedings{tan2018sim,
  title     = {Sim-to-Real: Learning Agile Locomotion For Quadruped Robots},
  author    = {Tan, Jie and Zhang, Tingnan and Coumans, Erwin and Iscen, Atil and Bai, Yunfei and Hafner, Danijar and Bohez, Steven and Vanhoucke, Vincent},
  booktitle = {Robotics: Science and Systems (RSS)},
  year      = {2018}
}

@inproceedings{chebotar2019closing,
  title        = {Closing the sim-to-real loop: Adapting simulation randomization with real world experience},
  author       = {Chebotar, Yevgen and Handa, Ankur and Makoviychuk, Viktor and Macklin, Miles and Issac, Jan and Ratliff, Nathan and Fox, Dieter},
  booktitle    = {2019 International Conference on Robotics and Automation (ICRA)},
  pages        = {8973--8979},
  year         = {2019},
  organization = {IEEE}
}

@inproceedings{deisenroth2011pilco,
  title     = {PILCO: A model-based and data-efficient approach to policy search},
  author    = {Deisenroth, Marc and Rasmussen, Carl E},
  booktitle = {Proceedings of the 28th International Conference on machine learning (ICML-11)},
  pages     = {465--472},
  year      = {2011}
}

@inproceedings{hafner2019learning,
  title        = {Learning latent dynamics for planning from pixels},
  author       = {Hafner, Danijar and Lillicrap, Timothy and Fischer, Ian and Villegas, Ruben and Ha, David and Lee, Honglak and Davidson, James},
  booktitle    = {International conference on machine learning},
  pages        = {2555--2565},
  year         = {2019},
  organization = {PMLR}
}

@inproceedings{hafner2020dream,
  title     = {Dream to Control: Learning Behaviors by Latent Imagination},
  author    = {Hafner, Danijar and Lillicrap, Timothy and Ba, Jimmy and Norouzi, Mohammad},
  booktitle = {International Conference on Learning Representations (ICLR)},
  year      = {2020}
}

@article{hafner2023mastering,
  title   = {Mastering diverse domains through world models},
  author  = {Hafner, Danijar and Pasukonis, Jurgis and Ba, Jimmy and Lillicrap, Timothy},
  journal = {arXiv preprint arXiv:2301.04104},
  year    = {2023}
}

@article{chua2018deep,
  title   = {Deep reinforcement learning in a handful of trials using probabilistic dynamics models},
  author  = {Chua, Kurtland and Calandra, Roberto and McAllister, Rowan and Levine, Sergey},
  journal = {Advances in neural information processing systems},
  volume  = {31},
  year    = {2018}
}

@inproceedings{wu2023daydreamer,
  title        = {Daydreamer: World models for physical robot learning},
  author       = {Wu, Philipp and Escontrela, Alejandro and Hafner, Danijar and Abbeel, Pieter and Goldberg, Ken},
  booktitle    = {Conference on robot learning},
  pages        = {2226--2240},
  year         = {2023},
  organization = {PMLR}
}

@inproceedings{golemo2018sim,
  title        = {Sim-to-real transfer with neural-augmented robot simulation},
  author       = {Golemo, Florian and Taiga, Adrien Ali and Courville, Aaron and Oudeyer, Pierre-Yves},
  booktitle    = {Conference on Robot Learning},
  pages        = {817--828},
  year         = {2018},
  organization = {PMLR}
}

@inproceedings{heiden2021neuralsim,
  title        = {NeuralSim: Augmenting differentiable simulators with neural networks},
  author       = {Heiden, Eric and Millard, David and Coumans, Erwin and Sheng, Yizhou and Sukhatme, Gaurav S},
  booktitle    = {2021 IEEE International Conference on Robotics and Automation (ICRA)},
  pages        = {9474--9481},
  year         = {2021},
  organization = {IEEE}
}

@article{hansen2023td,
  title   = {Td-mpc2: Scalable, robust world models for continuous control},
  author  = {Hansen, Nicklas and Su, Hao and Wang, Xiaolong},
  journal = {arXiv preprint arXiv:2310.16828},
  year    = {2023}
}

@article{gao2024sim,
  title     = {Sim-to-real of soft robots with learned residual physics},
  author    = {Gao, Junpeng and Michelis, Mike Y and Spielberg, Andrew and Katzschmann, Robert K},
  journal   = {IEEE Robotics and Automation Letters},
  year      = {2024},
  publisher = {IEEE}
}

@inproceedings{sontakke2023residual,
  title        = {Residual physics learning and system identification for sim-to-real transfer of policies on buoyancy assisted legged robots},
  author       = {Sontakke, Nitish and Chae, Hosik and Lee, Sangjoon and Huang, Tianle and Hong, Dennis W and Hal, Sehoon},
  booktitle    = {2023 IEEE/RSJ International Conference on Intelligent Robots and Systems (IROS)},
  pages        = {392--399},
  year         = {2023},
  organization = {IEEE}
}

@inproceedings{hanna2017grounded,
  title     = {Grounded action transformation for robot learning in simulation},
  author    = {Hanna, Josiah and Stone, Peter},
  booktitle = {Proceedings of the AAAI Conference on Artificial Intelligence},
  volume    = {31},
  number    = {1},
  year      = {2017}
}

@inproceedings{karnan2020rgat,
  title     = {Reinforced Grounded Action Transformation for Sim-to-Real Transfer},
  author    = {Karnan, Haresh and Desai, Siddharth and Hanna, Josiah P. and Warnell, Garrett and Stone, Peter},
  booktitle = {IEEE/RSJ International Conference on Intelligent Robots and Systems (IROS)},
  year      = {2020}
}

@inproceedings{he2025asap,
  title     = {ASAP: Aligning Simulation and Real-World Physics for Learning Agile Humanoid Whole-Body Skills},
  author    = {He, Tairan and Luo, Jia and He, Wenli and Xiao, Tong and Zhang, Yuzhe and Shi, Zhengyi and Fidler, Sanja and Zhu, Yuke and Liu, Linxi},
  booktitle = {Robotics: Science and Systems (RSS)},
  year      = {2025}
}

@inproceedings{zhu2017unpaired,
  title     = {Unpaired image-to-image translation using cycle-consistent adversarial networks},
  author    = {Zhu, Jun-Yan and Park, Taesung and Isola, Phillip and Efros, Alexei A},
  booktitle = {Proceedings of the IEEE international conference on computer vision},
  pages     = {2223--2232},
  year      = {2017}
}

@inproceedings{rao2020rl,
  title     = {Rl-cyclegan: Reinforcement learning aware simulation-to-real},
  author    = {Rao, Kanishka and Harris, Chris and Irpan, Alex and Levine, Sergey and Ibarz, Julian and Khansari, Mohi},
  booktitle = {Proceedings of the IEEE/CVF Conference on Computer Vision and Pattern Recognition},
  pages     = {11157--11166},
  year      = {2020}
}

@inproceedings{ho2021retinagan,
  title        = {Retinagan: An object-aware approach to sim-to-real transfer},
  author       = {Ho, Daniel and Rao, Kanishka and Xu, Zhuo and Jang, Eric and Khansari, Mohi and Bai, Yunfei},
  booktitle    = {2021 IEEE International Conference on Robotics and Automation (ICRA)},
  pages        = {10920--10926},
  year         = {2021},
  organization = {IEEE}
}

@article{mcinnes2018umap,
  title   = {Umap: Uniform manifold approximation and projection for dimension reduction},
  author  = {McInnes, Leland and Healy, John and Melville, James},
  journal = {arXiv preprint arXiv:1802.03426},
  year    = {2018}
}

@inproceedings{perez2018film,
  title     = {Film: Visual reasoning with a general conditioning layer},
  author    = {Perez, Ethan and Strub, Florian and De Vries, Harm and Dumoulin, Vincent and Courville, Aaron},
  booktitle = {Proceedings of the AAAI conference on artificial intelligence},
  volume    = {32},
  number    = {1},
  year      = {2018}
}

\end{document}